%% file: main_v3.tex
\documentclass{article}

\usepackage[preprint]{neurips_2026}
\usepackage[utf8]{inputenc} 
\usepackage[T1]{fontenc}    
\usepackage{hyperref}       
\usepackage{url}            
\usepackage{booktabs}       
\usepackage{amsfonts}       
\usepackage{nicefrac}       
\usepackage{microtype}      
\usepackage{xcolor}         

\usepackage[utf8]{inputenc}
\usepackage[T1]{fontenc}
\usepackage{hyperref}
\usepackage{url}
\usepackage{booktabs}
\usepackage{amsfonts}
\usepackage{nicefrac}
\usepackage{microtype}
\usepackage{xcolor}
\usepackage{amsmath}
\usepackage{amssymb}
\usepackage{enumitem}
\usepackage{pgfplots}
\pgfplotsset{compat=1.17}
\usepgfplotslibrary{groupplots}

\definecolor{cUniform}{HTML}{7F7F7F}
\definecolor{cRandom}{HTML}{BCBD22}
\definecolor{cFailure}{HTML}{9467BD}
\definecolor{cRisk}{HTML}{1F77B4}
\definecolor{cArm}{HTML}{17BECF}
\definecolor{cCtx}{HTML}{FF7F0E}
\definecolor{cOracle}{HTML}{2CA02C}

\pgfplotsset{
  paperaxis/.style={
    width=\linewidth, height=4.35cm,
    scale only axis,
    tick label style={font=\scriptsize},
    label style={font=\scriptsize},
    title style={font=\scriptsize\bfseries, yshift=-2pt},
    legend style={font=\scriptsize, draw=none, fill=none},
    grid=major, grid style={line width=0.25pt, draw=black!12},
    axis line style={line width=0.5pt, draw=black!55},
    xmin=0, xmax=730, xtick={0,100,200,300,500,700},
    every axis plot/.append style={line width=0.9pt, mark size=1.5pt},
    tick align=outside, tickpos=left,
  },
  sUniform/.style={color=cUniform, mark=square*, mark size=1.3pt},
  sRandom/.style={color=cRandom, mark=triangle*, mark size=1.8pt, densely dashed},
  sFailure/.style={color=cFailure, mark=diamond*, mark size=1.7pt},
  sRisk/.style={color=cRisk, mark=*},
  sArm/.style={color=cArm, mark=triangle*, mark size=1.8pt, densely dashed},
  sCtx/.style={color=cCtx, mark=*},
  sOracle/.style={color=cOracle, dashed, mark=none, line width=1.1pt},
}

\title{Risk-Aware Adaptive Evaluation: \\Finding High-Impact Failures Under Limited Budgets}

\author{%
  Priyanath Maji \\
  Georgia Institute of Technology\\
  Atlanta, GA 30332 \\
  \texttt{pmaji3@gatech.edu} \\
  \And
  Spandan Ghose Chowdhury \\
  Georgia Institute of Technology \\
  Atlanta, GA 30332 \\
  \texttt{spandan\_gc@gatech.edu} \\
}

\begin{document}

\maketitle

\begin{abstract}
Evaluating interactive agents is expensive. Agent behavior is stochastic, so reliability must be measured over repeated trials, but failures are rare and differ widely in how much they matter. Standard benchmarks spend this budget uniformly: a read-only lookup is sampled as often as an irreversible payment action. We instead formulate evaluation as a sequential allocation problem. Given a fixed trial budget and a set of scenarios whose failure behavior is unknown, which scenarios should be run, and run again? We propose a risk-aware contextual Thompson Sampling policy that combines a pre-execution scenario context vector and a fixed impact score with the failure outcomes observed during evaluation, and we test it by offline replay over 70 $\tau$-bench airline scenarios and 824 recorded trials. Our main result is at the smallest budget: with only 50 trials ($6\%$ of the corpus), the policy recovers $86\%$ of the impact-weighted failures an oracle could find, compared to $25\%$ for uniform allocation. It discovers $3.5\times$ more impact-weighted failures (215.4 vs.\ 62.2) with the same number of trials, delivers $5\times$ the discovery per dollar, and cuts the budget wasted on scenarios that never fail from $34\%$ to $2.8\%$. The rest of our analysis demonstrates and qualifies this result: a budget sweep shows the advantage shrinks as the budget approaches the corpus size, and paired significance tests show that scenario context helps mainly at small budgets while posterior-based exploration helps at moderate ones. Risk-aware adaptive allocation therefore helps most exactly where evaluation budget is scarcest.

\end{abstract}

\section{Introduction}
\label{sec:intro}

Large language model (LLM) agents increasingly perform multi-step
interactive tasks that involve tool use, changes to environment state, domain policy constraints, and direct interaction with users. Benchmarks such as $\tau$-bench~\cite{yao2024tau} and
$\tau^2$-bench~\cite{barres2025tau2} have shown that evaluating these agents requires more than checking a final answer: behavior must be judged over full trajectories, including tool execution, policy adherence, and state consistency. $\tau$-bench also shows that state-of-the-art function-calling agents are substantially inconsistent across repeated runs of the same scenario. Reliability, not single-run success, is the right object of measurement.

This view has a direct consequence for cost. If behavior is stochastic, one trial per scenario is a poor estimate of failure, so a benchmark must repeat trials. But repeating trials uniformly wastes budget, because scenarios differ along two independent axes. They differ in how often they fail: in our corpus, 24 of 70 scenarios never fail in any observed trial.
They also differ in what a failure means: a failed read-only lookup and a failed irreversible payment action are not equally important, yet uniform sampling treats them identically. As evaluation increasingly relies on repeated, costly trials against live tool-calling environments and LLM-based user simulators, this waste turns directly into evaluation cost.

Prior work on efficiency in agentic systems has mostly optimized the agent's computation rather than the evaluator's allocation of
trials, for example by optimizing multi-agent execution
graphs~\cite{zhuge2024gptswarm} or pruning redundant agents and
communication~\cite{wang2025agentdropout}. These methods reduce the cost of a single execution. We explore a complementary question: how many executions of which scenarios are worth running at all?

\paragraph{This work.} We frame interactive-agent evaluation as a
sequential resource-allocation problem. Given a fixed budget and a corpus of scenarios with unknown failure behavior, we examine which scenarios should be executed, and re-executed, to uncover as many \emph{consequential} failures as possible. We build on Thompson Sampling~\cite{thompson1933likelihood}, a Bayesian bandit strategy with well-understood regret behavior~\cite{agrawal2012analysis, russo2018tutorial}, extended with scenario-level context~\cite{agrawal2013thompson}. Allocation is then informed both by properties known \emph{before} a scenario is ever executed (whether it involves payment, irreversible actions, or policy constraints) and by outcomes observed \emph{during} evaluation. Unlike a standard bandit, the goal is not reward from an environment but the discovery of impact-weighted failures under a resource constraint: each scenario is a source of information about agent reliability.

\paragraph{Main result.} Our central finding concerns the small-budget regime: the setting where an agent is re-evaluated after every model, prompt, or policy change, so each decision can afford only a small slice of the corpus rather than many full passes \cite{huang2026howmany}. At a budget of 50 trials ($6\%$ of the corpus), contextual Thompson Sampling recovers $86.1\%$ of the impact-weighted failure discovery an oracle could achieve, against $24.9\%$ for uniform allocation. Paired, Holm-corrected tests over 30 shared-seed replicates decompose where this advantage comes from: pre-execution scenario context is what wins the cold start (beating the arm-only variant at $B \le 100$, $p_{\mathrm{Holm}} < 0.005$), while posterior-based exploration takes over at moderate budgets (beating the static risk-aware heuristic at $B = 300$--$500$, $p_{\mathrm{Holm}} < 10^{-8}$). Both effects, and the advantage itself, shrink as the budget grows toward the corpus size.

\paragraph{Contributions.} (1) \emph{Formulation}: we cast interactive-agent evaluation as a risk-aware sequential allocation problem that separates fixed, pre-execution quantities (context vector $\mathbf{x}_i$, impact score $I_i$) from stochastic per-trial observations (failure outcome $F_{i,t}$, resource use $\mathbf{R}_{i,t}$), with impact-weighted failure discovery as the objective. (2) \emph{Method}: we instantiate this as a contextual, risk-aware Thompson Sampling policy and compare it against uniform, random, failure-rate, and static risk-aware baselines and an oracle upper bound. (3) \emph{Analysis}: with paired, Holm-corrected tests we identify \emph{where} context and posterior-based exploration each help, not just whether they help. (4) \emph{Cost accounting}: we report discovery per dollar of measured evaluation cost, not just per trial.

\paragraph{Hypotheses.} We test four claims.
\textbf{H1 (Adaptivity)}: under a fixed budget, adaptive allocation discovers more impact-weighted failures than uniform allocation. \textbf{H2 (Context, budget-dependent)}: Thompson Sampling using observed outcomes, impact, and (in its contextual variant) pre-execution features $\mathbf{x}_i$ allocates more efficiently than non-adaptive allocation, though not necessarily at every budget. \textbf{H3 (Concentration)}: as trials accumulate, adaptive allocation increasingly avoids scenarios that show no evidence of failure while still exploring uncertain ones. \textbf{H4 (Cost efficiency)}: adaptive allocation discovers more impact-weighted failures per unit of measured cost, especially at small budgets.

\section{Related Work}
\label{sec:related}

\paragraph{Interactive agent benchmarks and reliability.}
AgentBench~\cite{liu2024agentbench} established the need to evaluate language models as agents in interactive environments rather than through static question answering. Later benchmarks extended this to web-based multi-step tasks~\cite{zhou2024webarena}, large-scale tool use over real APIs~\cite{qin2023toolllm}, and complex multi-step tool use over live MCP servers~\cite{wang2025mcpbench}. $\tau$-bench~\cite{yao2024tau} is closest to our setting: it evaluates tool-agent-user interaction under policy constraints and stateful environments, and its $\mathrm{Pass}^k$ metric measures reliability across repeated trials rather than single-run correctness; $\tau^2$-bench~\cite{barres2025tau2} extends this to dual-control environments. Follow-up studies confirm this variance is real and large: Mustahsan et al.~\cite{mustahsan2025stochasticity} quantify within-task inconsistency with intraclass correlation, Bjarnason et al.~\cite{bjarnason2026randomness} show by replay that single-run pass@1 on SWE-Bench-Verified moves by 2 to 6 points depending on which run is observed even at temperature 0, and Gonzalez-Pumariega et al.~\cite{gonzalez2026reliability} reach the same conclusion for computer-use agents. Other work enriches what each trajectory is scored on: TRAJECT-Bench~\cite{he2025traject} adds trajectory-level diagnostics beyond final success, and Cao et al.~\cite{cao2026beyond} show agents can reach correct final states through flawed procedures, so failures differ in kind and severity; we likewise extract operational signals (tool calls, turns, trajectory length, cost) from each trajectory.
All of these protocols, however, execute a predetermined, uniform trial schedule; even Huang~\cite{huang2026howmany}, who asks how many tasks a benchmark needs for a stable decision, works under uniform allocation. We formulate the schedule itself as the thing to optimize.

\paragraph{Sequential allocation and Thompson Sampling.}
Our problem is a multi-armed bandit: the evaluator picks a scenario, observes its outcome, and uses that to inform the next pick. Thompson Sampling~\cite{thompson1933likelihood} keeps a posterior over each arm's unknown parameter and samples from it, balancing exploration and exploitation; its finite-time regret is well understood~\cite{agrawal2012analysis, russo2018tutorial}, and Agrawal and Goyal~\cite{agrawal2013thompson} extend it to contextual bandits with linear payoffs, the mechanism we adapt for pre-execution features $\mathbf{x}_i$. Bastani et al.~\cite{bastani2017mostly} show that informative context can remove much of the need for exploration, consistent with our finding that context matters most in the cold start, and Riquelme et al.~\cite{riquelme2018deep} show that practical performance depends on the quality of the posterior approximation. Because our failure outcome is binary, the contextual version is most closely related to generalized linear bandits~\cite{filippi2010parametric}, fully Bayesian logistic Thompson Sampling via P\'olya-Gamma augmentation~\cite{dumitrascu2018pgts}, and neural variants~\cite{zhang2020neural}; extensions handle partially observable context~\cite{park2024partially} and parallel
pulls~\cite{karbasi2021parallelizing}. We use a simpler empirical-Bayes approximation, a lightweight choice at our corpus scale ($N=70$); a fully Bayesian generalized linear bandit is the natural choice at much larger scenario counts.

\paragraph{Budget-constrained discovery elsewhere.}
The problem of splitting a finite execution budget to find rare, costly failures is well studied outside agent evaluation. In software security, greybox fuzzers allocate executions across candidate test inputs to maximize bug discovery: Woo et al.~\cite{woo2013scheduling} first cast seed scheduling as a multi-armed bandit, and EcoFuzz~\cite{yue2020ecofuzz} models it adversarially to reduce wasted effort on well-explored inputs. In rare-event simulation and structural reliability, adaptive importance sampling and adaptive stratified sampling concentrate simulation effort in high-risk regions and avoid empty strata~\cite{geyer2019cross,
song2022adaptive, pieraccini2025adaptive}, and related work replaces uniform sampling with learned non-uniform sampling when many rare network threats must be simulated at once~\cite{liu2023rare}. None of these methods
study LLM agents, but they are the strongest precedent for our claim that rare, heterogeneous, high-impact failures call for adaptive allocation rather than uniform budgeting. The objective also matters: for estimating \emph{mean} accuracy under a fixed budget, spreading budget across more items with fewer trials each can be the variance-optimal choice~\cite{mustahsan2025stochasticity}. Our objective is different, discovery of impact-weighted failures, and that is where concentration pays. Two further things separate our setting from this work. First, we maximize discovered failures weighted by consequence, not environmental reward, an objective closer to active testing than to online decision-making. Second, each scenario's impact $I_i$ is known before execution while its failure outcome $F_{i,t}$ is observed only after, so our formulation separates a fixed notion of \emph{consequence} from an adaptively estimated notion of \emph{likelihood}, a distinction that does not arise when an arm's value is fully determined by its reward distribution.
\section{Method}
\label{sec:method}

\subsection{Problem Formulation}
\label{sec:formulation}

Let $\mathcal{S}=\{s_1,\dots,s_N\}$ be a set of evaluation scenarios. Each
scenario is a complete interactive task specification: a user objective, an
initial environment state, applicable policy constraints, and a ground-truth
reference action sequence. Each scenario carries a fixed context vector
$\mathbf{x}_i \in \mathbb{R}^{d}$, $d=17$, extracted \emph{statically from
the task definition} (its reference action sequence and natural-language
instruction), never from an observed outcome. This separation is deliberate:
$\mathbf{x}_i$ must be available before a scenario's first trial is
selected, and conditioning it on execution results would leak outcome
information into the allocation policy. The features are eight binary
task-structure indicators (\texttt{requires\_search}, 
\texttt{requires\_mutation}, \texttt{requires\_confirmation},
\texttt{has\_payment}, \texttt{date\_change}, \texttt{cabin\_change},
\texttt{multi\_segment}, \texttt{optimization\_required}), a passenger count
and single-passenger indicator, membership- and insurance-sensitivity
indicators, a complexity score and expected interaction horizon, an
irreversible-action indicator, a policy-constraint indicator, and a scalar
ambiguity score derived from instruction keywords. Three features
(\texttt{membership\_sensitive}, \texttt{complexity\_score},
\texttt{expected\_horizon}) are constant in this corpus; we keep them for
interface stability, and all variance-based analysis should be read against
the \emph{14 active features}.

Each trial $t$ of scenario $s_i$ produces a trajectory $\tau_{i,t} =
(s_0,a_1,o_1,\dots,a_T,o_T)$, where $a$ is an agent action or tool call and
$o$ the resulting observation. Because both the agent and the simulated user
are stochastic, repeated trials of the same scenario give different
trajectories and outcomes. From each trajectory we obtain a binary failure
indicator $F_{i,t}\in\{0,1\}$, determined by comparing the realized outcome
against the ground-truth reference, and a resource vector $\mathbf{R}_{i,t} =
[R^{\mathrm{tools}}_{i,t}, R^{\mathrm{turns}}_{i,t},
R^{\mathrm{traj}}_{i,t}, R^{\mathrm{cost}}_{i,t}]$ recording tool calls,
conversational turns, trajectory length, and measured monetary cost. Each
scenario also carries a fixed impact score $I_i$, giving the observed
failure-impact signal $Y_{i,t} = I_i F_{i,t}$. Repeated trials share
$\mathbf{x}_i$ and $I_i$ (both fixed and known before execution) while
$F_{i,t}$, $\mathbf{R}_{i,t}$, and $Y_{i,t}$ vary from trial to trial.

Given a trial budget $B$, the evaluator sequentially selects scenarios
$A_1,\dots,A_B$, observes $(F_{A_t,t}, \mathbf{R}_{A_t,t})$ after each
selection, and uses the history $\mathcal{H}_{t-1}$ to inform later choices,
$A_t \sim \pi(\mathcal{S}\mid\mathcal{H}_{t-1})$. The objective is
\begin{equation}
\label{eq:objective}
\max_{\pi}\ \mathbb{E}_{\pi}\!\left[\sum_{t=1}^{B} I_{A_t} F_{A_t,t}\right].
\end{equation}

\subsection{Impact Score}
\label{sec:impact}

The impact score is a fixed, rule-based function of a scenario's context
features. It reflects the \emph{consequence} of a failure, not its
likelihood:
\begin{equation}
\label{eq:impact}
I_i = \begin{cases}
5, & \text{irreversible action expected},\\
4, & \text{mutation \emph{and} (payment \emph{or} confirmation required)},\\
3, & \text{mutation, no payment or confirmation},\\
1, & \text{informational / read-only}.
\end{cases}
\end{equation}
$I_i$ is computed once per scenario from $\mathbf{x}_i$ and stays fixed
throughout evaluation; it is never updated from observed outcomes. This is a
deliberate simplification. A learned or trajectory-conditioned impact model
is a natural extension, but we do not implement it here. For scenario $i$
after $n_i$ trials we also track the empirical failure rate $\hat p_i =
n_i^{-1}\sum_t F_{i,t}$ and mean cost $\hat C_i = n_i^{-1}\sum_t
R^{\mathrm{cost}}_{i,t}$; $\hat C_i$ is used only after the fact, for
cost-normalized metrics.

\subsection{Allocation Strategies}
\label{sec:strategies}

We compare six policies plus an oracle upper bound. All adaptive policies
use either an $\varepsilon$-greedy or a posterior-sampling selection rule.
No policy normalizes its score by estimated cost $\hat C_i$, so cost enters
only as a reported outcome. Any cost efficiency we observe is therefore a byproduct of failure-aware selection, not an explicitly optimized objective.

\textbf{Uniform} visits scenarios in a fixed, once-shuffled round-robin order, cycling back to the start and skipping scenarios whose trials are exhausted. \textbf{Random} draws uniformly at random from scenarios with remaining trials, independent of history. \textbf{Failure-rate} selects $\arg\max_i \hat p_i$ with probability $1-\varepsilon$ ($\varepsilon=0.1$),
using an optimistic $\hat p_i = 1$ for $n_i=0$ so every scenario is tried at least once, and selects uniformly at random otherwise. \textbf{Risk-aware} has the same $\varepsilon$-greedy structure but ranks by $I_i\hat p_i$, so scenarios with higher \emph{known} impact are preferred even before any evidence is observed. It is the natural static-impact heuristic that
posterior-based exploration must beat. \textbf{Thompson Sampling (arm-only)} keeps each scenario as an independent Beta posterior $\theta_i\sim\mathrm{Beta}(\alpha_i,\beta_i)$ over
its failure probability, initialized $\alpha_i=\beta_i=1$ and updated as $(\alpha_i,\beta_i)\leftarrow(\alpha_i+F_{i,t},\ \beta_i+1-F_{i,t})$. At each step a sample $\tilde\theta_i$ is drawn for every available scenario and the maximizer of
\begin{equation}
\label{eq:ts-score}
V_i = I_i\,\tilde\theta_i
\end{equation}
is selected. This variant does not use $\mathbf{x}_i$. \textbf{Thompson Sampling (contextual)} is identical to the arm-only variant, except that the Beta parameters are informed by a population-level model of $\mathbf{x}_i$. After every pull a logistic regression is refit on all pooled $(\mathbf{x}_j, F_{j,\cdot})$ observations seen so far, giving a predicted failure probability $\hat p(\mathbf{x}_i)$ for \emph{every} scenario, including those never yet executed. This prediction is blended with a scenario's own evidence as pseudo-counts,
\begin{equation}
\label{eq:contextual-prior}
a_i = 1 + \lambda\,\hat p(\mathbf{x}_i) + \textstyle\sum_t F_{i,t},
\qquad
b_i = 1 + \lambda\,\bigl(1-\hat p(\mathbf{x}_i)\bigr) + n_i - \textstyle\sum_t F_{i,t},
\end{equation}
with $\tilde\theta_i\sim\mathrm{Beta}(a_i,b_i)$ and selection by
Eq.~\ref{eq:ts-score}. The context weight $\lambda$ sets how strong the
population-level prior is relative to a scenario's own evidence;
$\lambda=0$ recovers the arm-only variant exactly. We fixed $\lambda=2$ a priori as a weakly-informative default before running the sweep, and report it
throughout, with an ablation over $\lambda\in\{2,8,16,32\}$ in
Appendix~\ref{app:lambda}. Both Thompson Sampling variants select the first
trial of each replicate uniformly at random, since the flat prior at $t=0$
makes the posterior-sampled choice indistinguishable from random selection. \textbf{Oracle} sorts all trials by $Y_{i,t}$ in descending order and reports the cumulative sum of the top $B$ values for a given replicate's realized outcomes. This is the best any policy could achieve against that exact realized dataset at budget $B$, and it automatically respects each scenario's finite trial count because it draws from the same already-collected outcomes.

\subsection{Corpus and Offline Replay Protocol}
\label{sec:replay}

Scenarios come from the airline domain of $\tau$-bench~\cite{yao2024tau}:
$N=70$ scenarios covering reservation search, booking, cancellation, flight
modification, passenger and baggage changes, and payment operations. Trial
outcomes are generated with a GPT-4o~\cite{openai2024gpt4o} tool-calling agent (temperature 0.0)
against a GPT-4o simulated user under $\tau$-bench's native harness;
$F_{i,t}$ compares the agent's realized action sequence and final
environment state against the ground-truth reference. Trial counts per
scenario are not perfectly uniform (6 to 17), a consequence of aggregating
independently executed result batches; the corpus totals 824 trials.

Rather than executing fresh trials for each policy, we evaluate all
strategies by \emph{replaying} them against this single collected corpus.
For each of $M=30$ replicates, a scenario's observed trials are shuffled
with a replicate-specific seed and served without replacement as a policy
pulls that scenario. This is standard offline bandit evaluation, valid under
the assumption that trial order does not affect a scenario's underlying
outcome distribution. Critically, every policy in a sweep uses the identical
seed schedule ($\text{seed} = 10000s + r$), so replicate $r$ presents the
\emph{exact same} shuffled ordering to every policy; per-replicate outcomes
are directly comparable, which is what licenses the paired tests.

\subsection{Metrics and Statistical Procedure}
\label{sec:metrics}

All metrics are computed over $M=30$ replicates at budgets
$B\in\{50,100,150,200,300,500,700\}$. We report mean cumulative $Y$ and raw
failure count $F$; oracle-normalized efficiency
$\mathrm{pct\_of\_oracle}(\pi,B) = 100\,\bar Y_\pi(B)/Y^{*}(B)$, with
$Y^{*}(B)$ the oracle value computed per replicate and averaged;
cost-normalized efficiency $\mathrm{FDE}_\$ = \bar Y_\pi(B) /
\overline{\mathrm{cost}}_\pi(B)$ and its raw-failure analogue; and
\emph{trivial-scenario allocation}, the fraction of a policy's budget spent
on scenarios with $\hat p_i = 0$ across \emph{all} of their trials in the
full corpus (not just those the policy pulled), a direct test of whether
adaptive allocation avoids scenarios with no observed failure potential.

Because replicate $r$ presents an identical trial ordering to every policy,
we compare policies pairwise with a paired $t$-test on per-replicate
cumulative $Y$ at each budget, correcting within each family of tests using
the Holm step-down procedure~\cite{holm1979simple}. We apply this to compare
\textsc{ts-contextual} against \textsc{ts-arm-only} and
\textsc{risk-aware} (14 tests: 2 comparisons $\times$ 7 budgets), and to
compare \textsc{ts-contextual} using its full 17-feature context against a
13-feature variant with the four features that directly determine $I_i$
removed (7 tests). The second family isolates whether any contextual
advantage is just information already captured by the impact score.

\section{Results}
\label{sec:results}

We evaluate all six strategies on the full corpus of $N=70$ scenarios and
824 realized trials, replaying each policy against the same collected
outcomes for 30 independent replicates. Unless
noted, \textsc{ts-contextual} uses $\lambda=2$. Section~\ref{sec:results-main}
presents the main result of the paper, discovery at a small budget. The
remaining subsections are supporting analyses: they show how the advantage
changes with budget, which mechanism drives it and when, what it costs in
dollars, and where the saved budget comes from.

\subsection{Main Result: Discovery at a Small Budget}
\label{sec:results-main}

\begin{table}[t]
\centering
\caption{Mean cumulative impact-weighted failures ($Y$) and percentage of
oracle-achievable $Y$ recovered, at four representative budgets. The
$B{=}50$ column is the main result; larger budgets are shown for context.
Bold marks the best strategy at each budget.}
\label{tab:main-results}
\small
\setlength{\tabcolsep}{5pt}
\begin{tabular}{lcccccccc}
\toprule
& \multicolumn{2}{c}{$B=50$} & \multicolumn{2}{c}{$B=100$}
& \multicolumn{2}{c}{$B=300$} & \multicolumn{2}{c}{$B=700$} \\
\cmidrule(lr){2-3}\cmidrule(lr){4-5}\cmidrule(lr){6-7}\cmidrule(lr){8-9}
Strategy & $Y$ & \%oracle & $Y$ & \%oracle & $Y$ & \%oracle & $Y$ & \%oracle \\
\midrule
Uniform            & 62.2  & 24.9 & 124.1 & 26.4 & 372.3 & 37.0 & 858.0  & 81.2 \\
Random             & 59.9  & 23.9 & 123.7 & 26.3 & 367.2 & 36.5 & 870.5  & 82.4 \\
Failure-rate       & 166.2 & 66.5 & 357.2 & 76.0 & 708.3 & 70.4 & \textbf{1056.2} & \textbf{99.9} \\
Risk-aware         & 205.3 & 82.1 & \textbf{382.1} & \textbf{81.3} & 830.1 & 82.5 & \textbf{1056.2} & \textbf{99.9} \\
TS (arm-only)      & 204.5 & 81.8 & 365.1 & 77.7 & \textbf{908.4} & \textbf{90.3} & 1054.0 & 99.7 \\
TS (contextual)    & \textbf{215.4} & \textbf{86.1} & 374.7 & 79.7 & 906.2 & 90.1 & 1053.6 & 99.7 \\
\midrule
Oracle             & 250.0 & 100 & 470.0 & 100 & 1006.0 & 100 & 1057.0 & 100 \\
\bottomrule
\end{tabular}
\end{table}

\begin{figure}[t]
\centering
\begin{tikzpicture}
\begin{groupplot}[
  paperaxis, width=0.375\linewidth, height=3.4cm,
  group style={group size=2 by 1, horizontal sep=1.5cm},
  legend columns=4,
  legend style={/tikz/every even column/.append style={column sep=5pt}},
]
\nextgroupplot[
  legend to name=mainlegend,
  title={(a) Cumulative discovery},
  xlabel={Budget $B$ (trials)}, ylabel={Impact-weighted failures $Y$},
  ymin=0, ymax=1150, ytick={0,200,400,600,800,1000},
]
\addplot[sUniform] coordinates {(0,0)(50,62.2)(100,124.1)(150,185.6)(200,246.7)(300,372.3)(500,613.7)(700,858.0)};
\addlegendentry{uniform}
\addplot[sRandom] coordinates {(0,0)(50,59.9)(100,123.7)(150,184.8)(200,247.0)(300,367.2)(500,610.9)(700,870.5)};
\addlegendentry{random}
\addplot[sFailure] coordinates {(0,0)(50,166.2)(100,357.2)(150,476.6)(200,593.6)(300,708.3)(500,968.3)(700,1056.2)};
\addlegendentry{failure-rate}
\addplot[sRisk] coordinates {(0,0)(50,205.3)(100,382.1)(150,543.0)(200,676.3)(300,830.1)(500,966.7)(700,1056.2)};
\addlegendentry{risk-aware}
\addplot[sArm] coordinates {(0,0)(50,204.5)(100,365.1)(150,529.0)(200,679.3)(300,908.4)(500,1013.3)(700,1054.0)};
\addlegendentry{TS (arm-only)}
\addplot[sCtx] coordinates {(0,0)(50,215.4)(100,374.7)(150,533.8)(200,681.2)(300,906.2)(500,1012.5)(700,1053.6)};
\addlegendentry{TS (contextual)}
\addplot[sOracle] coordinates {(0,0)(50,250.0)(100,470.0)(150,670.0)(200,870.0)(300,1006.0)(500,1057.0)(700,1057.0)};
\addlegendentry{oracle}
\nextgroupplot[
  title={(b) Fraction of oracle recovered},
  xlabel={Budget $B$ (trials)}, ylabel={\% of oracle-achievable $Y$},
  ymin=15, ymax=108, ytick={20,40,60,80,100},
]
\addplot[sUniform] coordinates {(50,24.88)(100,26.41)(150,27.70)(200,28.36)(300,37.01)(500,58.06)(700,81.18)};
\addplot[sRandom] coordinates {(50,23.95)(100,26.31)(150,27.58)(200,28.39)(300,36.50)(500,57.80)(700,82.36)};
\addplot[sFailure] coordinates {(50,66.49)(100,76.00)(150,71.13)(200,68.23)(300,70.41)(500,91.61)(700,99.92)};
\addplot[sRisk] coordinates {(50,82.12)(100,81.29)(150,81.04)(200,77.74)(300,82.51)(500,91.45)(700,99.92)};
\addplot[sArm] coordinates {(50,81.79)(100,77.67)(150,78.96)(200,78.08)(300,90.29)(500,95.87)(700,99.72)};
\addplot[sCtx] coordinates {(50,86.15)(100,79.73)(150,79.68)(200,78.30)(300,90.08)(500,95.79)(700,99.68)};
\addplot[dotted, black!60, mark=none, line width=0.7pt] coordinates {(0,100)(730,100)};
\end{groupplot}
\end{tikzpicture}\\[2pt]
\ref{mainlegend}
\caption{Impact-weighted failure discovery as a function of evaluation
budget, for all six allocation strategies and the oracle upper bound.
\textbf{(a)} raw cumulative $Y$; \textbf{(b)} the same result normalized by
the oracle. The adaptive advantage is largest at the smallest budget and
disappears as the budget approaches the size of the corpus.}
\label{fig:discovery}
\end{figure}
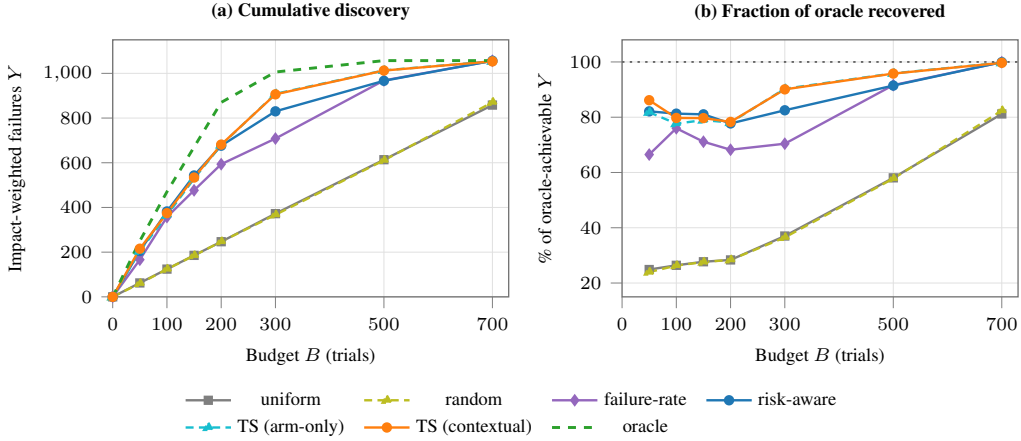

The main result of this paper is the $B=50$ column of
Table~\ref{tab:main-results}. At a budget of 50 trials, which is $6\%$ of
the 824-trial corpus, \textsc{ts-contextual} recovers $86.1\%$ of the
impact-weighted discovery an oracle could achieve. Uniform allocation
recovers $24.9\%$ and random allocation $23.9\%$. In raw terms,
\textsc{ts-contextual} finds $3.46\times$ more impact-weighted failures than
uniform (215.4 vs.\ 62.2) with the exact same number of trials. All four
adaptive strategies recover $66$--$86\%$ at this budget, supporting
\textbf{H1}: when the budget is tight, risk-aware adaptive allocation
finds far more of what matters.

We focus on the smallest budget because it is the realistic operating
point. The rest of the sweep
(Figure~\ref{fig:discovery}) bounds the claim: as the budget grows the
advantage shrinks, and by $B=700$ ($85\%$ of the corpus) all four adaptive
strategies converge to $99.7$--$99.9\%$ of the
oracle. Uniform and random remain at $81$--$82\%$ even at
that generous budget, because they keep spending trials on scenarios that
have already exhausted their contribution to $Y$. Adaptive allocation thus
dominates at every budget short of exhausting the corpus, but the size of
the advantage tracks how scarce the budget is. Nothing about the method is
specific to small budgets; small budgets are simply where it pays.

\subsection{Supporting Analysis: Which Mechanism Helps, and When}
\label{sec:results-stats}

\begin{table}[t]
\centering
\caption{Paired comparison of \textsc{ts-contextual} against
\textsc{ts-arm-only} (context vs.\ no context) and against
\textsc{risk-aware} (posterior exploration vs.\ a static impact-only
heuristic). $\Delta Y$ is the mean per-replicate difference; $p_{\mathrm{Holm}}$
is corrected across all 14 tests in this family. $^{*}$ marks
$p_{\mathrm{Holm}}<0.05$.}
\label{tab:pairwise}
\small
\begin{tabular}{lcccc}
\toprule
& \multicolumn{2}{c}{vs.\ \textsc{ts-arm-only}}
& \multicolumn{2}{c}{vs.\ \textsc{risk-aware}} \\
\cmidrule(lr){2-3}\cmidrule(lr){4-5}
Budget & $\Delta Y$ & $p_{\mathrm{Holm}}$ & $\Delta Y$ & $p_{\mathrm{Holm}}$ \\
\midrule
50  & $+10.9$ & $0.0001^{*}$ & $+10.1$ & $0.0002^{*}$ \\
100 & $+9.7$  & $0.0031^{*}$ & $-7.3$  & $0.075$ \\
150 & $+4.8$  & $0.543$      & $-9.2$  & $0.061$ \\
200 & $+1.9$  & $0.614$      & $+4.9$  & $0.398$ \\
300 & $-2.2$  & $0.543$      & $+76.1$ & $4.0\times10^{-9\,*}$ \\
500 & $-0.8$  & $0.614$      & $+45.8$ & $8.2\times10^{-10\,*}$ \\
700 & $-0.4$  & $0.543$      & $-2.6$  & $9.6\times10^{-7\,*}$ \\
\bottomrule
\end{tabular}
\end{table}

The main result shows that adaptive allocation wins at a small budget, but not which ingredient of \textsc{ts-contextual} is responsible.
Table~\ref{tab:pairwise} separates the two things it adds over blind allocation, scenario context $\mathbf{x}_i$ and posterior-based exploration, using the paired tests licensed by the shared seed schedule.

The comparison most relevant to the main result is the first row.
\emph{Context is what wins the cold start.} At $B\le100$,
\textsc{ts-contextual} beats \textsc{ts-arm-only}, which is identical except that it ignores $\mathbf{x}_i$, with $p_{\mathrm{Holm}}<0.005$. At no larger
budget is the difference distinguishable from noise. The explanation is simple: when almost nothing has been tried, the population-level model over
$\mathbf{x}_i$ is the only useful signal for unobserved scenarios, and its value fades as arms accumulate their own evidence.

A second effect appears at moderate budgets and is separate from context. \textsc{ts-contextual} beats the static heuristic \textsc{risk-aware} by a wide margin at $B=300$ and $B=500$ ($\Delta Y = +76.1$ and $+45.8$, $p_{\mathrm{Holm}}<10^{-8}$), even though it is indistinguishable from \textsc{ts-arm-only} at those same budgets. Since the two Thompson variants differ from \textsc{risk-aware} only in how they explore, this mid-budget advantage comes from posterior-based exploration, not from $\mathbf{x}_i$. \textsc{risk-aware} keeps exploring at a fixed rate ($\varepsilon=0.1$) no matter how much evidence it has, while the Thompson variants naturally explore less as their posteriors sharpen. We discover no advantage over either baseline at $B{=}150$--$200$. The nominally significant $B=700$ result against \textsc{risk-aware} is a $-2.6$ difference on 1056.2, a $0.24\%$ relative gap at a budget where the corpus is $85\%$ exhausted, and we do not read it as practically meaningful.

To summarize, the evidence for context is specific to the small-budget regime and the evidence for posterior exploration to the mid-budget regime; a blanket claim that contextual TS wins would be wrong in both directions. \textbf{H2} is therefore partially supported and explicitly budget-dependent.

Finally, we test whether the \emph{composition} of $\mathbf{x}_i$ matters,
comparing the full 17-feature vector against a residual 13-feature variant
with the four features that directly determine $I_i$ removed
(\texttt{requires\_mutation}, \texttt{requires\_confirmation},
\texttt{has\_payment}, \texttt{irreversible\_actions\_expected}). No budget
survives Holm correction ($p_{\mathrm{Holm}}\ge0.064$ everywhere;
Appendix~\ref{app:residual}). The contextual advantage is therefore not an
artifact of $\mathbf{x}_i$ re-encoding information already in $I_i$. At the
same time, the exact composition of the remaining features is not a
first-order driver of performance at this corpus size.

\subsection{Supporting Analysis: Cost in Dollars}
\label{sec:results-cost}

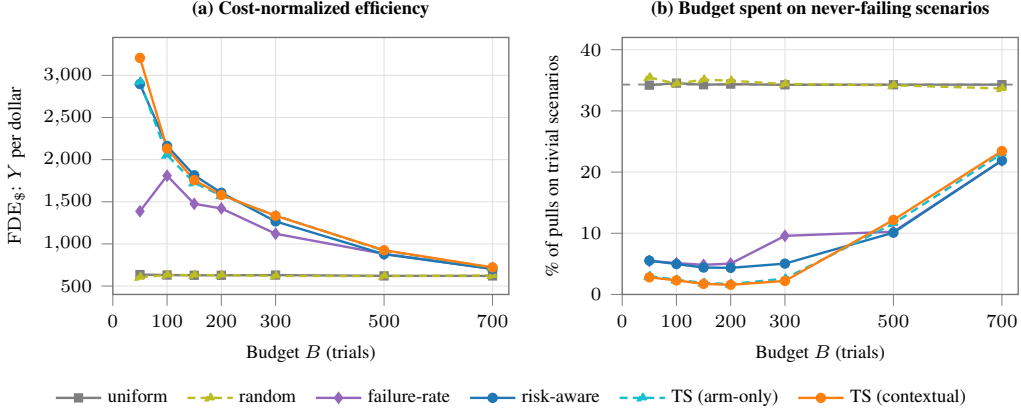
\begin{figure}[t]
\centering
\begin{tikzpicture}
\begin{groupplot}[
  paperaxis, width=0.375\linewidth, height=3.4cm,
  group style={group size=2 by 1, horizontal sep=1.5cm},
  legend columns=6,
  legend style={/tikz/every even column/.append style={column sep=5pt}},
]
\nextgroupplot[
  legend to name=costlegend,
  title={(a) Cost-normalized efficiency},
  xlabel={Budget $B$ (trials)}, ylabel={$\mathrm{FDE}_\$$: $Y$ per dollar},
  ymin=400, ymax=3450, ytick={500,1000,1500,2000,2500,3000},
]
\addplot[sUniform] coordinates {(50,635.9)(100,630.6)(150,627.8)(200,627.7)(300,629.9)(500,620.8)(700,620.2)};
\addplot[sRandom] coordinates {(50,605.1)(100,630.3)(150,631.1)(200,629.1)(300,620.2)(500,618.6)(700,625.5)};
\addplot[sFailure] coordinates {(50,1387.6)(100,1809.4)(150,1475.7)(200,1421.2)(300,1120.5)(500,881.4)(700,703.0)};
\addplot[sRisk] coordinates {(50,2896.6)(100,2162.1)(150,1814.2)(200,1606.4)(300,1267.1)(500,878.3)(700,702.9)};
\addplot[sArm] coordinates {(50,2924.7)(100,2053.5)(150,1725.1)(200,1572.7)(300,1338.2)(500,919.9)(700,717.8)};
\addplot[sCtx] coordinates {(50,3207.3)(100,2130.7)(150,1757.5)(200,1580.6)(300,1333.8)(500,925.2)(700,721.0)};
\legend{uniform, random, failure-rate, risk-aware, TS (arm-only), TS (contextual)}
\nextgroupplot[
  title={(b) Budget spent on never-failing scenarios},
  xlabel={Budget $B$ (trials)}, ylabel={\% of pulls on trivial scenarios},
  ymin=0, ymax=42, ytick={0,10,20,30,40},
]
\addplot[dashed, black!45, mark=none, line width=0.7pt] coordinates {(0,34.3)(730,34.3)};
\addplot[sUniform] coordinates {(50,34.20)(100,34.53)(150,34.27)(200,34.37)(300,34.26)(500,34.28)(700,34.29)};
\addplot[sRandom] coordinates {(50,35.47)(100,34.43)(150,35.11)(200,34.93)(300,34.39)(500,34.18)(700,33.66)};
\addplot[sFailure] coordinates {(50,5.47)(100,5.13)(150,4.84)(200,5.05)(300,9.58)(500,10.25)(700,21.89)};
\addplot[sRisk] coordinates {(50,5.53)(100,4.97)(150,4.40)(200,4.35)(300,5.04)(500,10.10)(700,21.90)};
\addplot[sArm] coordinates {(50,3.00)(100,2.40)(150,1.89)(200,1.70)(300,2.59)(500,11.58)(700,23.00)};
\addplot[sCtx] coordinates {(50,2.80)(100,2.30)(150,1.73)(200,1.58)(300,2.21)(500,12.18)(700,23.44)};
\end{groupplot}
\end{tikzpicture}\\[2pt]
\ref{costlegend}
\caption{\textbf{(a)} Impact-weighted failures discovered per dollar of
measured evaluation cost. \textbf{(b)} Fraction of budget spent on
\emph{trivial} scenarios, meaning those that never failed in any observed
trial (24 of 70; the grey dashed line is the $34.3\%$ corpus base rate).}
\label{fig:cost}
\end{figure}

\begin{table}[t]
\centering
\caption{Mean measured user-simulator cost (USD) and cost-normalized efficiency at the smallest and largest budgets. Cost is the logged \texttt{user\_cost} of the GPT-4o simulated user, the only component consistently recorded per trial; agent-side spend is not included. $Y/\$$ is $\mathrm{FDE}_\$$; $F/\$$ is its
raw-failure analogue.}
\label{tab:cost}
\small
\setlength{\tabcolsep}{6pt}
\begin{tabular}{lcccccc}
\toprule
& \multicolumn{3}{c}{$B=50$} & \multicolumn{3}{c}{$B=700$} \\
\cmidrule(lr){2-4}\cmidrule(lr){5-7}
Strategy & Cost (\$) & $Y/\$$ & $F/\$$ & Cost (\$) & $Y/\$$ & $F/\$$ \\
\midrule
Uniform         & 0.098 & 636  & 221 & 1.383 & 620 & 211 \\
Random          & 0.099 & 605  & 208 & 1.392 & 625 & 211 \\
Failure-rate    & 0.120 & 1388 & 347 & 1.502 & 703 & 233 \\
Risk-aware      & 0.071 & 2897 & 591 & 1.503 & 703 & 233 \\
TS (arm-only)   & 0.070 & 2925 & 591 & 1.468 & 718 & 237 \\
TS (contextual) & \textbf{0.067} & \textbf{3207} & \textbf{650} & 1.461 & \textbf{721} & \textbf{238} \\
\bottomrule
\end{tabular}
\end{table}

Trial count is a convenient proxy for budget, but the resource actually
spent is money. Table~\ref{tab:cost} and Figure~\ref{fig:cost}(a) restate
the main result in dollars. At $B=50$, \textsc{ts-contextual} finds
$5.0\times$ more impact-weighted failures per dollar than uniform allocation
(3207 vs.\ 636) while spending \emph{less} in total (\$0.067 vs.\ \$0.098).
The gain is not bought with extra spend; it comes from putting the same or
smaller outlay on more informative scenarios. Consistent with the
budget-dependence above, the advantage narrows to $1.16\times$ by $B=700$,
supporting \textbf{H4} at small budgets with diminishing, but never
negative, returns at large ones. This accounting uses directly logged
\texttt{user\_cost}. No policy selects on a cost-adjusted utility, so the observed cost efficiency is a
byproduct of failure-aware selection and a conservative estimate of what
cost-aware selection could achieve.

\subsection{Supporting Analysis: Where the Saved Budget Comes From}
\label{sec:results-trivial}

What does uniform allocation spend its budget on that
\textsc{ts-contextual} does not? To answer this, and to test \textbf{H3},
we label a scenario \emph{trivial} if it never failed
in any of its trials in the full corpus ($\hat p_i = 0$). Of the 70
scenarios, 24 ($34.3\%$) are trivial, so a policy that ignores observed
outcomes should spend about a third of its budget learning nothing.
Figure~\ref{fig:cost}(b) shows exactly that: uniform and random track the
$34.3\%$ base rate at every budget, within a percentage point. At the
main-result budget of $B=50$, \textsc{ts-contextual} spends $2.8\%$ on
trivial scenarios against uniform's $34.2\%$, a $91.8\%$ relative reduction.
All four adaptive strategies hold trivial spend below $6\%$ through
$B=200$ (\textsc{failure-rate} rises to $9.6\%$ at $B=300$; the other
three stay at $2$--$5\%$), and \textsc{ts-contextual} has the lowest trivial spend of all six
strategies at every budget through $B=300$. This is where the small-budget
advantage comes from: the trials that uniform allocation wastes on scenarios
that cannot fail are redirected to scenarios that can.

Trivial spend rises again for all adaptive strategies beyond $B=300$,
reaching $21$--$24\%$ by $B=700$. This is not a policy failure; it is what a
finite corpus forces. Once a strategy has used up the available trials in
every scenario capable of producing $Y>0$, the only budget left to spend is
on scenarios already known to be safe. That turning point is a useful diagnostic: it marks
the budget at which a fixed corpus has been mined of its discoverable
failures, beyond which extra budget buys confidence rather than discovery.

\subsection{Synthesis}
\label{sec:synthesis}

Across the four hypotheses: \textbf{H1} is supported (adaptive strategies
recover $66$--$86\%$ of oracle at $B=50$ vs.\ $23.9$--$24.9\%$ for
uniform/random); \textbf{H2} is partially supported and budget-dependent, in
the decomposed sense of Section~\ref{sec:results-stats}; \textbf{H3} is
supported (trivial spend falls from a $\sim$$34\%$ base rate to below $6\%$
through $B=200$, and to $2$--$5\%$ through $B=300$ for all but \textsc{failure-rate}); and \textbf{H4} is supported at low budgets with
diminishing returns at high ones ($\mathrm{FDE}_\$$ improves $5.0\times$ at
$B=50$, $1.16\times$ at $B=700$).

The practical reading: with a budget small relative to the corpus, the
common case, use contextual Thompson Sampling; with a budget near the
corpus size, any adaptive policy will do, including the much simpler
risk-aware heuristic. One caveat: three of our 17 features carry no
variance in this corpus, so the
residual-feature null speaks to the 14 active features, not to the full
intended feature space.

\section{Limitations}
\label{sec:limitations}

Four limitations bound these conclusions. \textbf{Scope}: results come from a single interactive airline-agent environment with 70 scenarios. Failure patterns in other domains such as customer support, healthcare, finance, or software engineering may differ, and replication across benchmarks is the most important next step. \textbf{Impact specification}: $I_i$ is rule-based and fixed; it cannot infer the severity of a new failure from its trajectory, and a learned impact model would be more principled. \textbf{Arm independence}: modeling each scenario as an independent arm ignores correlations. Scenarios sharing a policy or operation type likely share failure modes, which a hierarchical formulation would exploit. Relatedly, our contextual variant uses a plug-in empirical-Bayes approximation rather than a full posterior over the logistic link, a simplification more likely to matter at larger corpus sizes. \textbf{Objective}: we focus primarily on failure discovery under a fixed evaluation budget. Higher failure discovery does not necessarily imply greater failure diversity, severity, representativeness, or diagnostic value. Finally, our resource model uses directly measurable execution quantities (tool calls, turns, trajectory length, dollar cost); token-level consumption and wall-clock latency were not consistently available in these logs. These results are evidence that resource- and risk-aware evaluation is feasible and valuable, not a claim of universal optimality.

\section{Conclusion}
\label{sec:conclusion}

We formulated interactive-agent evaluation as a sequential resource allocation problem in which the evaluator, not the benchmark designer, decides how many trials each scenario deserves. Separating a scenario's fixed, pre-execution consequence ($I_i$, $\mathbf{x}_i$) from its observed likelihood of failure ($F_{i,t}$) yields a risk-aware contextual Thompson Sampling policy whose value is concentrated where budgets are tight. With $6\%$ of the corpus's trial budget, it recovers $86\%$ of what a perfectly informed oracle could find: $3.5\times$ what uniform allocation finds with the same trials, at $5\times$ the discovery per dollar, while cutting wasted effort on never-failing scenarios by an order of magnitude. Paired, Holm-corrected significance testing further shows that the benefit of scenario context specifically is concentrated in the cold-start regime and posterior exploration helps at moderate budgets. The broader point is simple: the trial schedule of an agent benchmark is a design variable worth optimizing, and treating it as fixed leaves a large share of a small evaluation budget on the table.

\paragraph{Future directions.} The most immediate extensions are a learned, trajectory-conditioned impact model in place of the rule-based $I_i$; a hierarchical or fully Bayesian generalized linear bandit that shares statistical strength across related scenarios; and cost-adjusted selection, since our policies achieve their cost efficiency without optimizing for it. Beyond these, replicating on more benchmarks and extending the objective toward failure diversity and calibrated reliability estimation would test how far the formulation generalizes.


\paragraph{Broad impact statement.} This work aims to make interactive-agent evaluation more efficient and more attentive to consequential failures under limited testing budgets. We see this as broadly beneficial for the safe deployment of agentic systems: prioritizing evaluation effort toward scenarios involving irreversible actions, payment, or policy-sensitive operations is intended to surface higher-stakes failures earlier and with less computational and monetary cost than uniform testing. However, an impact score that is misspecified or incomplete, could cause an adaptive policy to systematically under-sample consequential scenarios. Also, adaptive evaluation is a tool for finding failures faster, not a substitute for comprehensive evaluation coverage; a scenario that receives few trials because it currently shows no evidence of failure should not be interpreted as verified safe, only as unexamined by this allocation policy at this budget.

{\small
\bibliographystyle{unsrt}
\bibliography{references}
}

\newpage
\appendix
\section*{Appendix}

\section{Feature-Composition Ablation}
\label{app:residual}

Table~\ref{tab:full-vs-residual} compares \textsc{ts-contextual} using the
full 17-feature context vector against a residual 13-feature variant with
the four features that directly determine $I_i$ removed. No budget survives
Holm correction across this 7-test family, so the contextual advantage
reported in Section~\ref{sec:results-stats} is not explained by redundancy
with the impact score.

\begin{table}[h]
\centering
\caption{\textsc{ts-contextual} with the full 17-feature $\mathbf{x}_i$
versus a 13-feature residual vector with impact-determining features
removed. $\Delta$ is computed on unrounded per-replicate values, so it can differ from the difference of the rounded columns by $0.1$. $p_{\mathrm{Holm}}$ is corrected across all 7 tests.}
\label{tab:full-vs-residual}
\small
\begin{tabular}{lccccc}
\toprule
Budget & $Y_{\mathrm{full}}$ & $Y_{\mathrm{residual}}$ & $\Delta$ & $\mathrm{sd}(\Delta)$ & $p_{\mathrm{Holm}}$ \\
\midrule
50  & 215.4 & 213.9 & $+1.4$ & 7.58  & 1.000 \\
100 & 374.7 & 375.6 & $-0.9$ & 12.62 & 1.000 \\
150 & 533.8 & 534.5 & $-0.7$ & 12.46 & 1.000 \\
200 & 681.2 & 676.0 & $+5.3$ & 10.16 & 0.064 \\
300 & 906.2 & 906.7 & $-0.5$ & 6.30  & 1.000 \\
500 & 1012.5 & 1011.3 & $+1.2$ & 3.10 & 0.242 \\
700 & 1053.6 & 1054.2 & $-0.6$ & 1.64 & 0.242 \\
\bottomrule
\end{tabular}
\end{table}

\section{Full Trivial-Scenario Allocation Sweep}
\label{app:trivial}

\begin{table}[h]
\centering
\caption{Fraction (\%) of evaluation budget spent on trivial
(never-failing) scenarios, by strategy and budget. 24 of 70 scenarios
($34.3\%$) are trivial by this definition.}
\label{tab:trivial}
\small
\begin{tabular}{lccccccc}
\toprule
Strategy & $B{=}50$ & 100 & 150 & 200 & 300 & 500 & 700 \\
\midrule
Uniform         & 34.2 & 34.5 & 34.3 & 34.4 & 34.3 & 34.3 & 34.3 \\
Random          & 35.5 & 34.4 & 35.1 & 34.9 & 34.4 & 34.2 & 33.7 \\
Failure-rate    & 5.5  & 5.1  & 4.8  & 5.0  & 9.6  & 10.3 & 21.9 \\
Risk-aware      & 5.5  & 5.0  & 4.4  & 4.4  & 5.0  & 10.1 & 21.9 \\
TS (arm-only)   & 3.0  & 2.4  & 1.9  & 1.7  & 2.6  & 11.6 & 23.0 \\
TS (contextual) & \textbf{2.8} & \textbf{2.3} & \textbf{1.7} & \textbf{1.6} & \textbf{2.2} & 12.2 & 23.4 \\
\bottomrule
\end{tabular}
\end{table}

\section{Per-Scenario Allocation}
\label{app:heatmap}

\begin{figure}[htbp]
\centering
\includegraphics[width=0.9\linewidth]{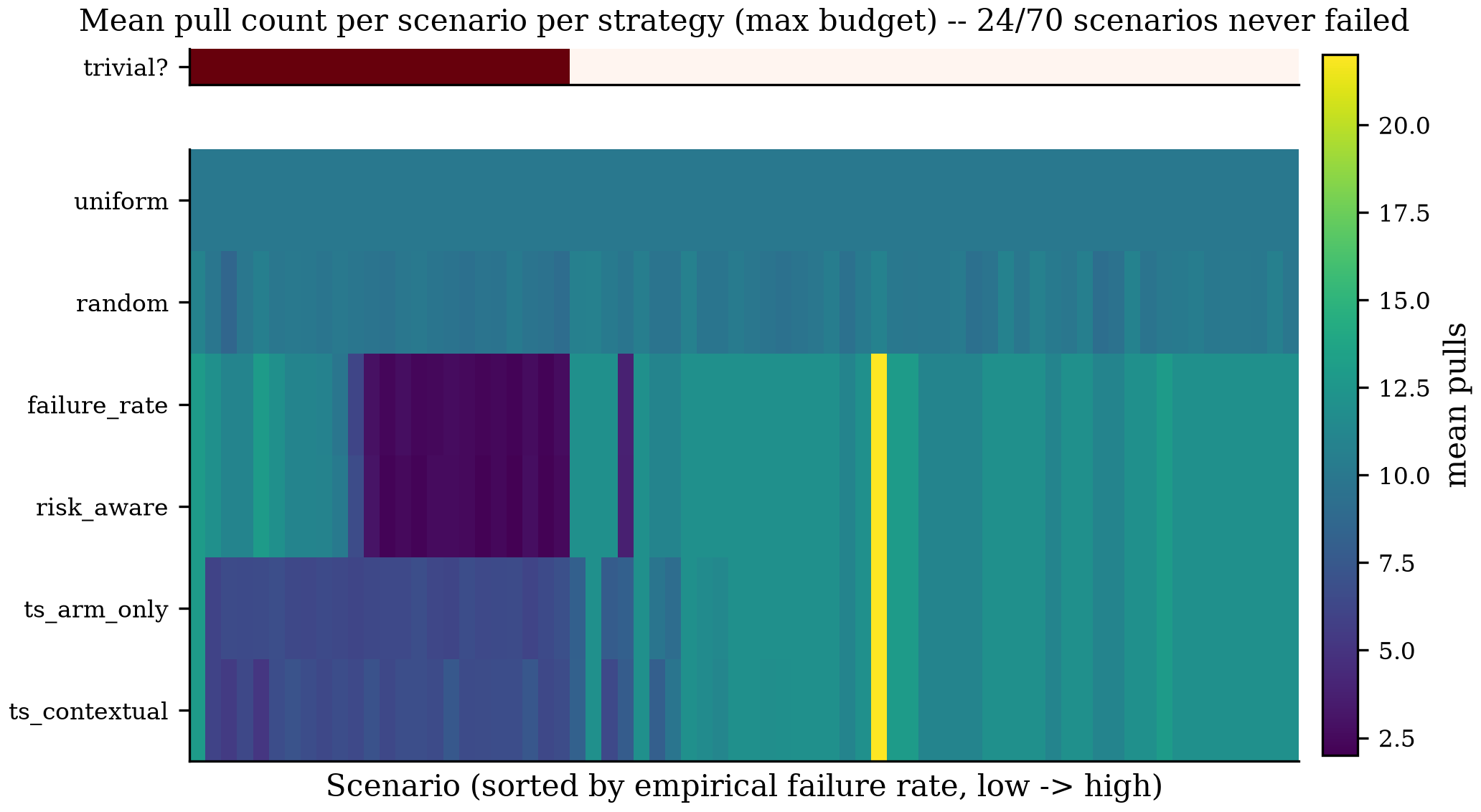}
\caption{Mean number of pulls allocated to each scenario at maximum budget,
by strategy, with scenarios sorted by empirical failure rate (low to high).
The red strip marks the 24 scenarios that never failed in the corpus.
Uniform and random allocate flatly across the sort order; the adaptive
strategies visibly withdraw from the never-failing block on the left.}
\label{fig:heatmap}
\end{figure}

Figure~\ref{fig:heatmap} shows how each strategy distributes its pulls
across individual scenarios at maximum budget. It makes the concentration
effect of Section~\ref{sec:results-trivial} visible at the level of single
scenarios rather than aggregate fractions.

\section{Context-Weight Ablation}
\label{app:lambda}

\begin{figure}[htbp]
\centering
\includegraphics[width=0.62\linewidth]{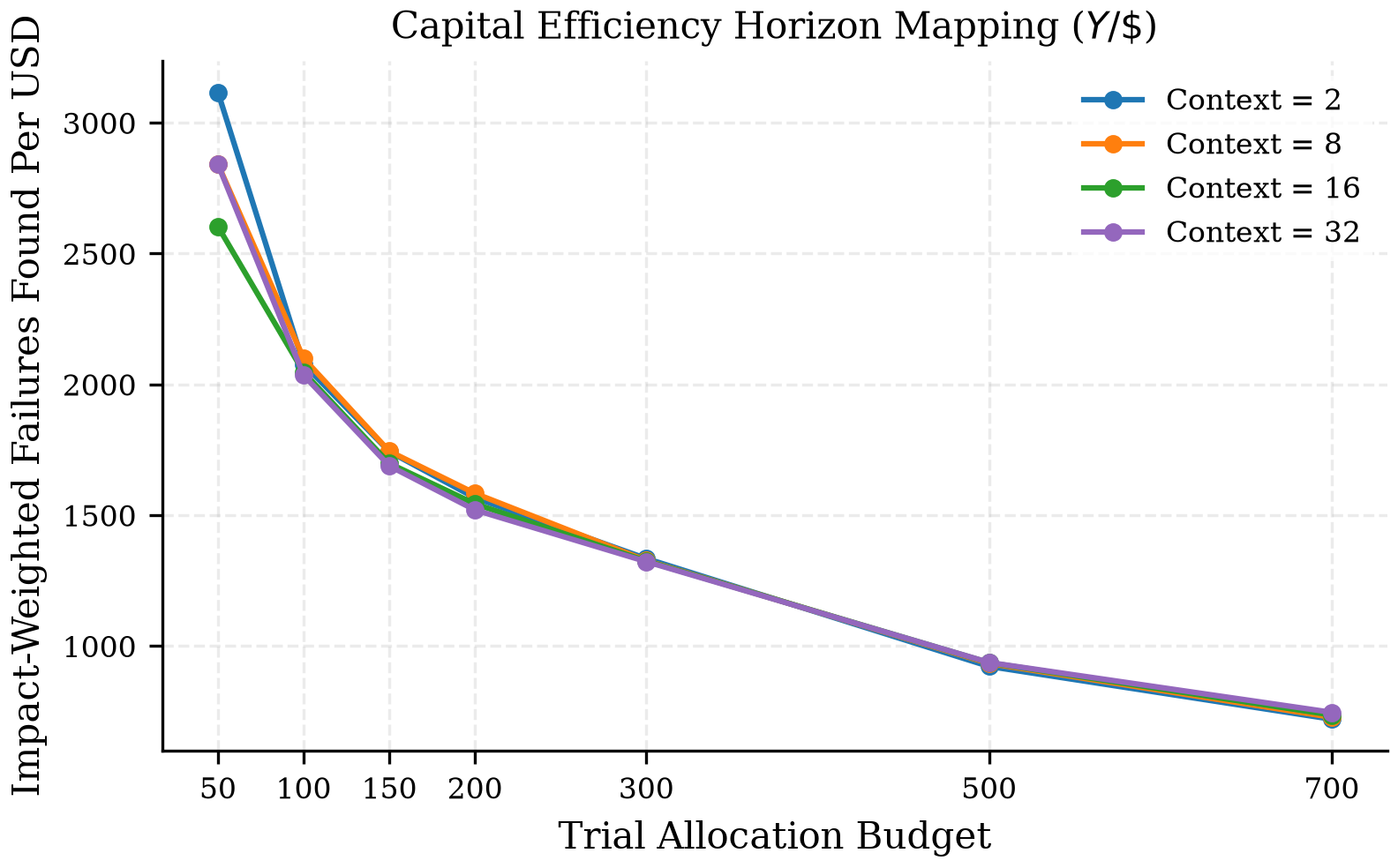}
\caption{Cost-normalized efficiency ($Y$ per dollar) for
\textsc{ts-contextual} across context weights $\lambda\in\{2,8,16,32\}$.}
\label{fig:lambda-eff}
\end{figure}

\begin{figure}[htbp]
\centering
\includegraphics[width=0.62\linewidth]{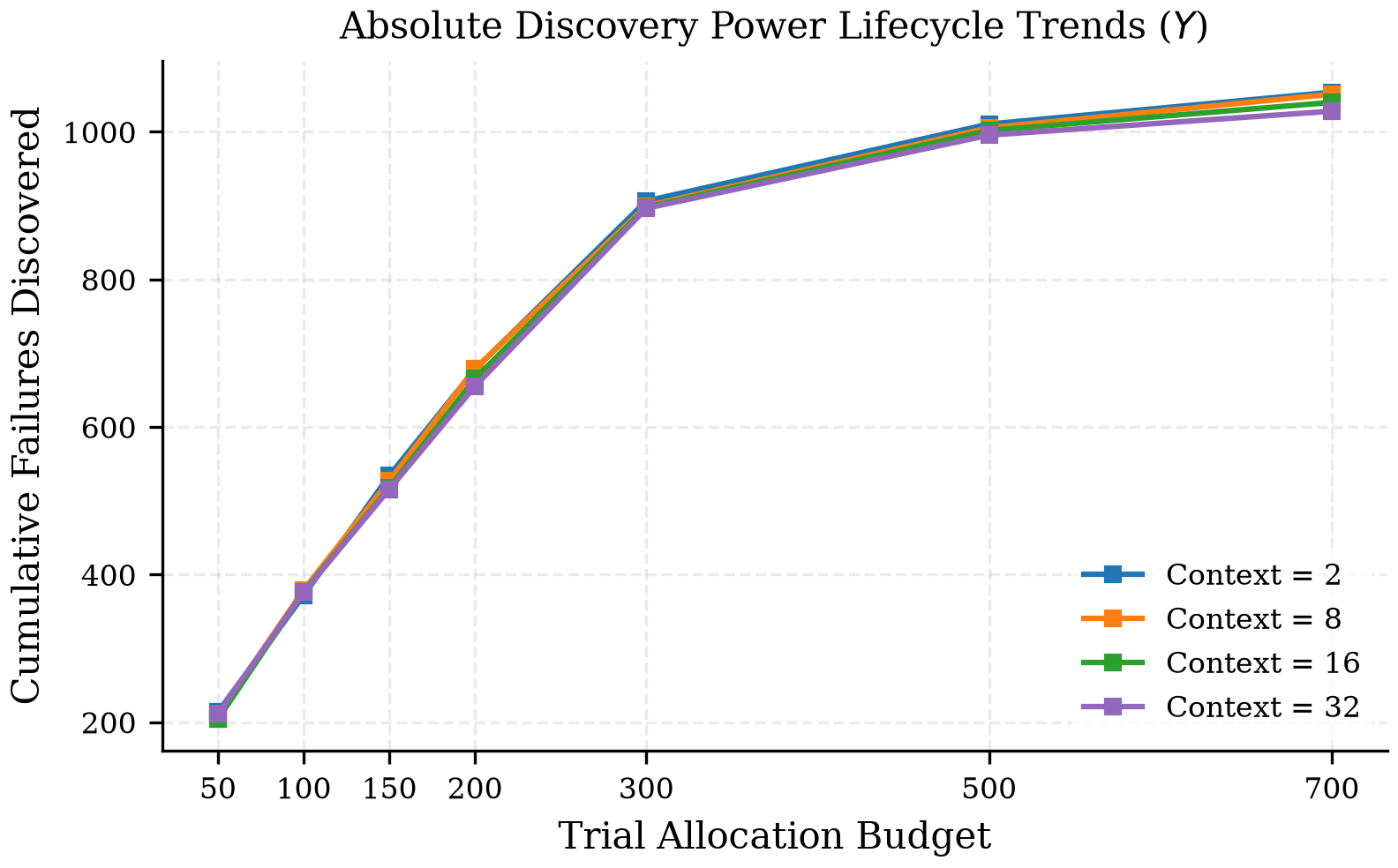}
\caption{Cumulative impact-weighted discovery for \textsc{ts-contextual}
across context weights $\lambda\in\{2,8,16,32\}$.}
\label{fig:lambda-yield}
\end{figure}

Figures~\ref{fig:lambda-eff} and~\ref{fig:lambda-yield} sweep the context
weight $\lambda\in\{2,8,16,32\}$ of Eq.~\ref{eq:contextual-prior}. The
choice of $\lambda$ matters only at the smallest budget, where the
population-level prior carries the most weight relative to per-arm evidence.
By $B\ge100$ the four settings are indistinguishable. We report
$\lambda = 2$ throughout the main text.

\section{Compute Resources}
\label{app:compute}
 
\paragraph{Corpus collection.} The 824 trials of Section~\ref{sec:replay} were
collected by running a GPT-4o tool-calling agent against a GPT-4o simulated user
through the $\tau$-bench harness.
 
\paragraph{Replay experiments.} Every result in Sections~\ref{sec:results-main}
through~\ref{sec:results-trivial} and in Appendices~\ref{app:residual}
and~\ref{app:lambda} is produced by offline replay against the fixed corpus and
involves no model inference. The full sweep including six policies plus the oracle,
seven budgets, 30 replicates, plus the residual-feature and $\lambda$ ablations runs in approximately 15 minutes on NVIDIA H200, 80 GB of memory.
 
\section{Assets and Licenses}
\label{app:assets}
 
\paragraph{$\tau$-bench.} Scenarios, the environment, the reference action
sequences, and the user simulator come from the airline domain of
$\tau$-bench~\cite{yao2024tau} (version 59a200c6d575d595120f1cb70fea53cef0632f6b), released by its authors under the MIT License. We use the 50 scenarios from the benchmark unmodified for trial execution and added 20 custom scenarios for outcome scoring; the allocation layer studied here sits above it and changes only which scenarios are run and how often.
 
\paragraph{GPT-4o.} Both the evaluated agent and the simulated user are GPT-4o
(gpt-4o-2025-03-01), accessed through the OpenAI API under
the OpenAI Terms of Use and API data-usage policy. No model weights were obtained
or redistributed.
 
\paragraph{Software.} Analysis and replay use Python 3.12 with NumPy, SciPy,
scikit-learn (logistic regression for the contextual prior), pandas, and
Matplotlib, tau-bench, each under its own permissive open-source license (BSD-3-Clause or
equivalent).
 
\paragraph{Released artifacts.} The replay corpus and the allocation and replay
code released with this paper are made available under the MIT
License. The corpus is a derived record of agent behavior on $\tau$-bench and
contains no data beyond per-trial outcomes, resource counts, and measured cost;
it is redistributed in a form consistent with the upstream MIT license, which is
reproduced in the repository.

\newpage
\newpage

\input{checklist.tex}
\end{document}

%% file: checklist.tex
\section*{NeurIPS Paper Checklist}

\begin{enumerate}

\item {\bf Claims}
    \item[] Question: Do the main claims made in the abstract and introduction accurately reflect the paper's contributions and scope?
    \item[] Answer: \answerYes{}
    \item[] Justification: The abstract and Section~1 state exactly what the paper delivers: a risk-aware sequential-allocation formulation of interactive-agent evaluation, a contextual Thompson Sampling policy, and the headline small-budget result ($86.1\%$ of oracle-achievable impact-weighted discovery at $B{=}50$ versus $24.9\%$ for uniform allocation). Both also state the scope limits we verify in Sections~4.2 and 4.5: the advantage is budget-dependent and vanishes as the budget approaches the corpus size, and hypothesis H2 is reported as only partially supported.

\item {\bf Limitations}
    \item[] Question: Does the paper discuss the limitations of the work performed by the authors?
    \item[] Answer: \answerYes{}
    \item[] Justification: Section~5 is a dedicated Limitations section covering the single-domain scope (70 airline scenarios from one benchmark), the fixed rule-based impact score, the independent-arm assumption together with the plug-in empirical-Bayes approximation used in place of a full posterior over the logistic link, the discovery-only objective, and the incomplete resource model. Section~4.5 additionally flags that three of the seventeen context features carry no variance in this corpus, so the residual-feature null of Appendix~A speaks to the fourteen active features.

\item {\bf Theory assumptions and proofs}
    \item[] Question: For each theoretical result, does the paper provide the full set of assumptions and a complete (and correct) proof?
    \item[] Answer: \answerNA{}
    \item[] Justification: The paper contains no theorems, lemmas, or formal claims requiring proof; the contribution is a problem formulation together with an empirical study. The regret properties of Thompson Sampling that motivate the method are cited \cite{thompson1933likelihood, agrawal2012analysis, russo2018tutorial, agrawal2013thompson} rather than re-derived here.

    \item {\bf Experimental result reproducibility}
    \item[] Question: Does the paper fully disclose all the information needed to reproduce the main experimental results of the paper to the extent that it affects the main claims and/or conclusions of the paper (regardless of whether the code and data are provided or not)?
    \item[] Answer: \answerYes{}
    \item[] Justification: Section~3.1 specifies the 17-dimensional context vector feature by feature, Equation~2 gives the impact rule in full, and Section~3.3 defines all six allocation policies and the oracle with every hyperparameter stated ($\varepsilon = 0.1$, $\lambda = 2$, $\mathrm{Beta}(1,1)$ initialization, and the pseudo-count blending of Equation~4). Section~3.4 gives the corpus and the offline replay protocol including the exact seed schedule ($\text{seed} = 10000s + r$ over $M = 30$ replicates) that makes every reported number regenerable, and Section~3.5 defines each metric and the paired-testing procedure.

\item {\bf Open access to data and code}
    \item[] Question: Does the paper provide open access to the data and code, with sufficient instructions to faithfully reproduce the main experimental results, as described in supplemental material?
    \item[] Answer: \answerYes{}
    \item[] Justification: An repository containing the allocation policies, the offline replay harness, the 824-trial corpus, and scripts that regenerate every table and figure in the paper is provided with the submission (\url{https://github.com/priyanathmaji/risk-aware-eval/}); its README gives the exact commands and the Python environment. Because all reported results come from replay against a fixed collected corpus rather than fresh model calls, reproducing them requires no API credentials or paid inference.

\item {\bf Experimental setting/details}
    \item[] Question: Does the paper specify all the training and test details (e.g., data splits, hyperparameters, how they were chosen, type of optimizer) necessary to understand the results?
    \item[] Answer: \answerYes{}
    \item[] Justification: Section~3.3 states every policy hyperparameter, Section~3.4 describes the corpus and the replay protocol (including that per-scenario trial counts are non-uniform, 6 to 17, as a consequence of aggregating independently executed batches), and Section~3.5 lists the budgets $B \in \{50, 100, 150, 200, 300, 500, 700\}$ and the $M = 30$ replicates. There is no model training; the only free parameter is the context weight $\lambda$, fixed a priori at $2$ and swept over $\{2, 8, 16, 32\}$ in Appendix~D.

\item {\bf Experiment statistical significance}
    \item[] Question: Does the paper report error bars suitably and correctly defined or other appropriate information about the statistical significance of the experiments?
    \item[] Answer: \answerYes{}
    \item[] Justification: Every reported number is a mean over $M = 30$ replicates that differ only in the seed controlling trial-order shuffling, which is the sole source of randomness under replay. Because all policies are served an identical seed schedule, we compare them with paired $t$-tests on per-replicate cumulative $Y$ at each budget and correct with the Holm step-down procedure within each family of tests (14 tests in Table~2, 7 in Table~4, which also reports $\mathrm{sd}(\Delta)$); the procedure is specified in Section~3.5.

\item {\bf Experiments compute resources}
    \item[] Question: For each experiment, does the paper provide sufficient information on the computer resources (type of compute workers, memory, time of execution) needed to reproduce the experiments?
    \item[] Answer: \answerYes{}
    \item[] Justification: Appendix~E reports the hardware used, the wall-clock time and total OpenAI API spend required to collect the 824-trial corpus (including preliminary and discarded runs), and the negligible CPU cost of the replay sweeps themselves. Table~3 additionally reports the logged per-trial user-simulator cost that enters the cost-normalized metric.
    
\item {\bf Code of ethics}
    \item[] Question: Does the research conducted in the paper conform, in every respect, with the NeurIPS Code of Ethics \url{https://neurips.cc/public/EthicsGuidelines}?
    \item[] Answer: \answerYes{}
    \item[] Justification: We have reviewed the NeurIPS Code of Ethics and the research conforms to it in every respect. The study involves no human subjects and no personal or sensitive data, since both the agent and the user are language models acting in the synthetic $\tau$-bench airline environment; the submission is anonymized and the released artifacts carry no dual-use risk.

\item {\bf Broader impacts}
    \item[] Question: Does the paper discuss both potential positive societal impacts and negative societal impacts of the work performed?
    \item[] Answer: \answerYes{}
    \item[] Justification: The Broad Impact Statement in Section~6 discusses both directions. Cheaper evaluation lowers the barrier to testing deployed agents thoroughly and weighting discovery by consequence directs scarce effort toward high-stakes failures; against that, concentrating budget on high-impact scenarios under-samples the rest and weakens reliability estimates for low-priority scenarios, and the impact score $I_i$ encodes a value judgement about which failures count. We state the mitigation explicitly: adaptive allocation is not a substitute for uniform coverage when the goal is an unbiased reliability estimate, and the two objectives should be budgeted separately.
    
\item {\bf Safeguards}
    \item[] Question: Does the paper describe safeguards that have been put in place for responsible release of data or models that have a high risk for misuse (e.g., pre-trained language models, image generators, or scraped datasets)?
    \item[] Answer: \answerNA{}
    \item[] Justification: The paper releases no pretrained models, generative systems, or scraped data. The released artifacts are evaluation code and replay logs of agent outcomes on a public synthetic benchmark, which carry no meaningful risk of misuse.

\item {\bf Licenses for existing assets}
    \item[] Question: Are the creators or original owners of assets (e.g., code, data, models), used in the paper, properly credited and are the license and terms of use explicitly mentioned and properly respected?
    \item[] Answer: \answerYes{}
    \item[] Justification: Appendix~F credits every external asset and states its version and license: the $\tau$-bench airline domain \cite{yao2024tau} under its MIT license, GPT-4o accessed through the OpenAI API under the OpenAI terms of use, and the open-source Python libraries used for the analysis. All assets are used within their stated terms.

\item {\bf New assets}
    \item[] Question: Are new assets introduced in the paper well documented and is the documentation provided alongside the assets?
    \item[] Answer: \answerYes{}
    \item[] Justification: The submission introduces two assets, both in the anonymized repository referenced in item~5: the 824-trial replay corpus (per-trial failure indicator, resource vector, and measured cost for each of the 70 scenarios) and the allocation and replay implementation. Both ship with a README documenting the record schema, the collection procedure of Section~3.4, and the license.

\item {\bf Crowdsourcing and research with human subjects}
    \item[] Question: For crowdsourcing experiments and research with human subjects, does the paper include the full text of instructions given to participants and screenshots, if applicable, as well as details about compensation (if any)? 
    \item[] Answer: \answerNA{}
    \item[] Justification: The study involves no human participants. The ``user'' in every trial is a GPT-4o simulator supplied by the $\tau$-bench harness, as described in Section~3.4.

\item {\bf Institutional review board (IRB) approvals or equivalent for research with human subjects}
    \item[] Question: Does the paper describe potential risks incurred by study participants, whether such risks were disclosed to the subjects, and whether Institutional Review Board (IRB) approvals (or an equivalent approval/review based on the requirements of your country or institution) were obtained?
    \item[] Answer: \answerNA{}
    \item[] Justification: No human subjects were involved in this research, so no IRB review or equivalent approval was required.

\item {\bf Declaration of LLM usage}
    \item[] Question: Does the paper describe the usage of LLMs if it is an important, original, or non-standard component of the core methods in this research? Note that if the LLM is used only for writing, editing, or formatting purposes and does \emph{not} impact the core methodology, scientific rigor, or originality of the research, declaration is not required.
    \item[] Answer: \answerYes{}
    \item[] Justification: GPT-4o is a core component of the experimental pipeline rather than a writing aid: it serves both as the tool-calling agent under evaluation (temperature $0.0$) and as the simulated user that drives each interaction, and the 824 trial outcomes replayed throughout the paper are its output (Section~3.4). The allocation method itself including the impact rule, the Beta posteriors, and the contextual prior uses no LLM.

\end{enumerate}

%% file: references.bib
@inproceedings{yao2024tau,
  title={$\tau$-bench: A Benchmark for Tool-Agent-User Interaction in Real-World Domains},
  author={Yao, Shunyu and Shinn, Noah and Razavi, Pedram and Narasimhan, Karthik},
  booktitle={International Conference on Learning Representations},
  volume={2025},
  pages={9965--10017},
  year={2025}
}

@inproceedings{
barres2025tau2,
title={$\tau^2$-Bench: Evaluating Conversational Agents in a Dual-Control Environment},
author={Victor Barres and Honghua Dong and Soham Ray and Xujie Si and Karthik R Narasimhan},
booktitle={Forty-third International Conference on Machine Learning},
year={2026},
url={https://openreview.net/forum?id=OC2z7iSQKa}
}

@article{agrawal2012analysis,
  title={Analysis of Thompson Sampling for the Multi-armed Bandit Problem},
  author={Agrawal, Shipra and Goyal, Navin},
  journal={arXiv preprint arXiv:1111.1797},
  year={2012}
}

@inproceedings{agrawal2013thompson,
  title={Thompson Sampling for Contextual Bandits with Linear Payoffs},
  author={Agrawal, Shipra and Goyal, Navin},
  booktitle={International Conference on Machine Learning},
  pages={127--135},
  year={2013}
}

@inproceedings{zhuge2024gptswarm,
  title={GPTSwarm: Language Agents as Optimizable Graphs},
  author={Zhuge, Mingchen and Wang, Wenyi and Kirsch, Louis and Faccio, Francesco and Khizbullin, Dmitrii and Schmidhuber, J{\"u}rgen},
  booktitle={Proceedings of the 41st International Conference on Machine Learning},
  volume={235},
  pages={62743--62767},
  year={2024}
}

@inproceedings{wang2025agentdropout,
  title={AgentDropout: Dynamic Agent Elimination for Token-Efficient and High-Performance LLM-Based Multi-Agent Collaboration},
  author={Wang, Zhexuan and Wang, Yutong and Liu, Xuebo and Ding, Liang and Zhang, Miao and Liu, Jie and Zhang, Min},
  booktitle={Proceedings of the 63rd Annual Meeting of the Association for Computational Linguistics},
  pages={24013--24035},
  year={2025}
}

@inproceedings{liu2024agentbench,
    title={AgentBench: Evaluating LLMs as Agents},
    author={Liu, Xiao and Yu, Hao and Zhang, Hanchen and Xu, Yifan and Lei, Xuanyu and Lai, Hanyu and Gu, Yu and Ding, Hangliang and Men, Kaiwen and Yang, Kejuan and Zhang, Shudan and Deng, Xiang and Zeng, Aohan and Du, Zhengxiao and Zhang, Chenhui and Shen, Sheng and Su, Tianjun and Sun, Huan and Huang, Minlie and Dong, Yuxiao and Tang, Jie},
    booktitle={International Conference on Learning Representations},
    year={2024}
}

@inproceedings{zhou2024webarena,
    title={WebArena: A Realistic Web Environment for Building Autonomous Agents},
    author={Zhou, Shuyan and Xu, Frank F. and Zhu, Hao and Zhou, Xuhui and Lo, Robert and Sridhar, Abishek and Cheng, Xueqiao and Bisk, Yonatan and Fried, Daniel and Alon, Uri and Roth, Dan},
    booktitle={International Conference on Learning Representations},
    year={2024}
}

@inproceedings{qin2023toolllm,
    title={ToolLLM: Facilitating Large Language Models to Master 16000+ Real-world APIs},
    author={Qin, Yujia and Liang, Shihao and Ye, Yining and Zhu, Kunlun and Yan, Lan and Lu, Yaxi and Lin, Yankai and Cong, Xue and Tang, Xiangru and Qian, Bill and others},
    booktitle={International Conference on Learning Representations},
    year={2024}
}

@article{thompson1933likelihood,
    title={On the Likelihood that One Unknown Probability Exceeds Another
           in View of the Evidence of Two Samples},
    author={Thompson, William R.},
    journal={Biometrika},
    volume={25},
    number={3--4},
    pages={285--294},
    year={1933}
}

@book{russo2018tutorial,
    title={A Tutorial on Thompson Sampling},
    author={Russo, Daniel and Van Roy, Benjamin and Kazerouni, Abbas and
            Osband, Ian and Wen, Zheng},
    journal={Foundations and Trends in Machine Learning},
    volume={11},
    number={1},
    pages={1--96},
    year={2018}
}

@article{holm1979simple,
  title={A simple sequentially rejective multiple test procedure},
  author={Holm, Sture},
  journal={Scandinavian Journal of Statistics},
  volume={6},
  number={2},
  pages={65--70},
  year={1979},
  publisher={JSTOR}
}

@inproceedings{filippi2010parametric,
 author = {Filippi, Sarah and Cappe, Olivier and Garivier, Aur\'{e}lien and Szepesv\'{a}ri, Csaba},
 booktitle = {Advances in Neural Information Processing Systems},
 editor = {J. Lafferty and C. Williams and J. Shawe-Taylor and R. Zemel and A. Culotta},
 pages = {},
 publisher = {Curran Associates, Inc.},
 title = {Parametric Bandits: The Generalized Linear Case},
 url = {https://proceedings.neurips.cc/paper_files/paper/2010/file/c2626d850c80ea07e7511bbae4c76f4b-Paper.pdf},
 volume = {23},
 year = {2010}
}

@inproceedings{dumitrascu2018pgts,
 author = {Dumitrascu, Bianca and Feng, Karen and Engelhardt, Barbara},
 booktitle = {Advances in Neural Information Processing Systems},
 editor = {S. Bengio and H. Wallach and H. Larochelle and K. Grauman and N. Cuester-Blanchet and R. Garnett},
 pages = {7713--7723},
 publisher = {Curran Associates, Inc.},
 title = {PG-TS: Improved Thompson Sampling for Logistic Contextual Bandits},
 url = {https://neurips.cc},
 volume = {31},
 year = {2018}
}

@inproceedings{woo2013scheduling,
author = {Woo, Maverick and Cha, Sang Kil and Gottlieb, Samantha and Brumley, David},
title = {Scheduling black-box mutational fuzzing},
year = {2013},
isbn = {9781450324779},
publisher = {Association for Computing Machinery},
address = {New York, NY, USA},
url = {https://doi.org/10.1145/2508859.2516736},
doi = {10.1145/2508859.2516736},
booktitle = {Proceedings of the 2013 ACM SIGSAC Conference on Computer \& Communications Security},
pages = {511–522},
numpages = {12},
location = {Berlin, Germany},
series = {CCS '13}
}

@inproceedings{yue2020ecofuzz,
author = {Yue, Tai and Wang, Pengfei and Tang, Yong and Wang, Enze and Yu, Bo and Lu, Kai and Zhou, Xu},
title = {EcoFuzz: adaptive energy-saving greybox fuzzing as a variant of the adversarial multi-armed bandit},
year = {2020},
isbn = {978-1-939133-17-5},
publisher = {USENIX Association},
address = {USA},
articleno = {130},
numpages = {18},
series = {SEC'20}
}

@article{mustahsan2025stochasticity,
  title={Stochasticity in Agentic Evaluations: Quantifying Inconsistency with Intraclass Correlation},
  author={Mustahsan, Zairah and Lim, Abel and Anand, Megna and Jain, Saahil and McCann, Bryan T},
  journal={arXiv preprint arXiv:2512.06710},
  year={2025}
}

@article{bjarnason2026randomness,
  title={On Randomness in Agentic Evals},
  author={Bjarnason, Bjarni Haukur and Silva, Andr{\'e} and Monperrus, Martin},
  journal={arXiv preprint arXiv:2602.07150},
  year={2026}
}

@article{gonzalez2026reliability,
  title={On the Reliability of Computer Use Agents},
  author={Gonzalez-Pumariega, Gonzalo and Agashe, Saaket and Yang, Jiachen and Li, Ang and Wang, X.},
  journal={arXiv preprint arXiv:2604.17849},
  year={2026}
}

@article{cao2026beyond,
  title={Beyond Task Completion: Revealing Corrupt Success in {LLM} Agents through Procedure-Aware Evaluation},
  author={Cao, Hongliu and Driouich, Ilias and Thomas, Eoin},
  journal={arXiv preprint arXiv:2603.03116},
  year={2026}
}

@inproceedings{
he2025traject,
title={{TRAJECT}-Bench:A Trajectory-Aware Benchmark for Evaluating Agentic Tool Use},
author={Pengfei He and Zhenwei Dai and Bing He and Hui Liu and Xianfeng Tang and Hanqing Lu and Juanhui Li and Jiayuan Ding and Subhabrata Mukherjee and Suhang Wang and Yue Xing and Jiliang Tang and Benoit Dumoulin},
booktitle={The Fourteenth International Conference on Learning Representations},
year={2026},
url={https://openreview.net/forum?id=TZWnWvsQ0X}
}

@article{wang2025mcpbench,
  title={{MCP-Bench}: Benchmarking Tool-Using {LLM} Agents with Complex Real-World Tasks via {MCP} Servers},
  author={Wang, Zhenting and Chang, Qi and Patel, Hemani and Biju, S. and Wu, Chen and Liu, Quan and Ding, Aolin and Rezazadeh, Alireza and Shah, Ankit and Bao, Yujia and Siow, Eugene},
  journal={arXiv preprint arXiv:2508.20453},
  year={2025}
}

@article{huang2026howmany,
  title={How Many Tasks Are Enough for Agent Benchmark Decisions? {A} Replay Analysis of Public {LLM} Agent Benchmarks},
  author={Huang, Wei-Jung},
  journal={arXiv preprint arXiv:2607.12338},
  year={2026}
}

@article{bastani2017mostly,
  title={Mostly Exploration-Free Algorithms for Contextual Bandits},
  author={Bastani, Hamsa and Bayati, Mohsen and Khosravi, Khashayar},
  journal={Management Science},
  volume={67},
  pages={1329--1349},
  year={2021}
}

@article{riquelme2018deep,
  title={Deep {Bayesian} Bandits Showdown: An Empirical Comparison of {Bayesian} Deep Networks for {Thompson} Sampling},
  author={Riquelme, Carlos and Tucker, George and Snoek, Jasper},
  journal={arXiv preprint arXiv:1802.09127},
  year={2018}
}

@article{zhang2020neural,
  title={Neural {Thompson} Sampling},
  author={Zhang, Weitong and Zhou, Dongruo and Li, Lihong and Gu, Quanquan},
  journal={arXiv preprint arXiv:2010.00827},
  year={2020}
}

@article{park2024partially,
  title={{Thompson} Sampling in Partially Observable Contextual Bandits},
  author={Park, Hongju and Faradonbeh, Mohamad Kazem Shirani},
  journal={arXiv preprint arXiv:2402.10289},
  year={2024}
}

@inproceedings{karbasi2021parallelizing,
  title={Parallelizing {Thompson} Sampling},
  author={Karbasi, Amin and Mirrokni, Vahab and Shadravan, Mohammad},
  booktitle={Advances in Neural Information Processing Systems},
  pages={10535--10548},
  year={2021}
}

@article{pieraccini2025adaptive,
  title={An adaptive importance sampling algorithm for risk-averse optimization},
  author={Pieraccini, S. and Vanzan, Tommaso},
  journal={arXiv preprint arXiv:2502.10084},
  year={2025}
}

@article{song2022adaptive,
  title={Adaptive stratified sampling for structural reliability analysis},
  author={Song, Chenxiao and Kawai, Reiichiro},
  journal={Structural Safety},
  volume={101},
  pages={102292},
  year={2022}
}

@article{geyer2019cross,
  title={Cross entropy-based importance sampling using {Gaussian} densities revisited},
  author={Geyer, S. and Papaioannou, I. and Straub, D.},
  journal={Structural Safety},
  volume={76},
  pages={15--27},
  year={2019}
}

@article{liu2023rare,
  title={Online Zero-Cost Learning: Optimizing Large Scale Network Rare Threats Simulation},
  author={Liu, Tingwei and Xie, Hong and Lui, John C.S.},
  journal={IEEE Transactions on Mobile Computing},
  volume={22},
  pages={356--373},
  year={2023}
}

@misc{openai2024gpt4o,
  author       = {{OpenAI}},
  title        = {{ChatGPT} ({GPT-4o} version)},
  year         = {2024},
  note         = {Large language model Accessed: 2026-07-29 - 2026-08-29},
  howpublished = {Computer software},
  url          = {https://chatgpt.com},
}
